\documentclass{amia}
\usepackage{graphicx}
\usepackage[superscript,nomove]{cite}
\usepackage{color}
\usepackage{caption} 
\usepackage{multirow} 

\usepackage[
singlelinecheck=false 
]{caption}
\usepackage{titlesec}
\titlespacing*{\section}{0pt}{*0}{*0}
\usepackage{subfiles}
\newif\ifsubfile
\subfiletrue

\usepackage{soul}
\usepackage[color=yellow, textsize=small]{todonotes}
\usepackage{booktabs}
\usepackage[hidelinks]{hyperref}
\usepackage{hyperref}
\usepackage{url}
\usepackage{enumitem}
\makeatletter
\renewcommand{\@biblabel}[1]{\hfill #1.}
\makeatother
\usepackage{graphicx}
\usepackage[export]{adjustbox}
\usepackage{wrapfig,lipsum,booktabs}

\let\oldbibliography\thebibliography
\renewcommand{\thebibliography}[1]{%
  \oldbibliography{#1}%
  \setlength{\itemsep}{-0.1em}%
}

\usepackage{multicol}
\usepackage{calc} 

\title{Adapting Biomedical Abstracts into Plain language using Large Language Models}

\author{ 
\begin{center}
Haritha Gangavarapu$^2$\footnote{Equal contribution}, Giridhar Kaushik Ramachandran$^1$\footnotemark[1], Kevin Lybarger$^1$, \\
Meliha Yetisgen$^3$, Özlem Uzuner$^1$
\end{center}
}

\institutes{ 
\begin{center}
    $^1$Department of Information Sciences \& Technology, George Mason University, Fairfax, VA \\ $^2$CGI inc.  $^3$University of Washington, Seattle, WA, USA 
\end{center}
}

\begin{document}

\maketitle

\subfilefalse
\begin{abstract}

A vast amount of medical knowledge is available for public use through online health forums, and question-answering platforms on social media. The majority of the population in the United States doesn't have the right amount of health literacy to make the best use of that information. Health literacy means the ability to obtain and comprehend the basic health information to make appropriate health decisions. To build the bridge between this gap, organizations advocate adapting this medical knowledge into plain language. Building robust systems to automate the adaptations helps both medical and non-medical professionals best leverage the available information online. The goal of the Plain Language Adaptation of Biomedical Abstracts (PLABA) track is to adapt the biomedical abstracts in English language extracted from PubMed based on the questions asked in MedlinePlus for the general public using plain language at the sentence level. As part of this track, we leveraged the best open-source Large Language Models suitable and fine-tuned for dialog use cases. We compare and present the results for all of our systems and our ranking among the other participants' submissions. Our top performing GPT-4 based model ranked first in the avg. simplicity measure and 3\textsuperscript{rd} on the avg. accuracy measure.

\ifsubfile
\bibliography{mybib}
\fi

\end{abstract}
\setcounter{page}{7}
\section{Introduction}

The TAC-2023 Plain Language Adaptation of Biomedical Abstracts (PLABA)\cite{PLABA-TAC-url,attal2022dataset} focused on adapting technical language in biomedical abstracts into plain language for the general public use. The PLABA dataset is curated from sources like MedlinePlus and PubMed. MedlinePlus is an online health information resource by the National Library of Medicine (NLM), that contains the relevant health and wellness information available and accessible for free for medical professionals as well as the general public. PubMed supports the search and retrieval of biomedical literature. Due to the availability of vast information from these reliable resources, the public can easily access and search for any health-related question to get the answers they are seeking. However, the content available may not always be suitable for the general public with limited or no medical background as these sources contain information from scientific journals which may not be easily understood without relevant education and prior knowledge.

Studies \cite{Berkman2011, Pourhabibi2022} have analyzed the impact of poor health literacy on health outcomes. So organizations like Centers for Disease Control (CDC), and  (NIH) maintain materials and resources that are helpful for the public to understand the research in health and also promote plain language adaptation by providing guidelines for effective adaptation\cite{CDC-plain-language, NIH-plain-language}. Prior research on the readability of Patient Education Materials highlights the need to simplify the available biomedical information\cite{Jasem2022, Eltorai2022, Tulbert2011, Kher2012, PEM2012} to promote effective self-disease management\cite{Pourhabibi2022, Marciano2019} and awareness to identify existing medical conditions\cite{Rogers2006}.

In this paper, we explore automatically adapting biomedical abstracts into plain language using Large Language Models. We use the Large Language models with parameter efficient fine-tuning as well as In Context Learning approach, to make the models adapt to the dataset by learning from the analogy. In our work, we compare the different versions of GPT-J and LLaMa second generation models. Our system ranked 1\textsuperscript{st} in the external manual evaluation on the overall average simplicity measure and 3\textsuperscript{rd} on the average accuracy measure.

\ifsubfile
\bibliography{mybib}
\fi

\section{Related Work}


There has been a great level of advocacy for plain language adaptation in the biomedical domain because of its helpfulness to the public. Research has been conducted to assess the availability of plain-language summaries\cite{FitzGibbon2020} and explore ways to improve consumer engagement\cite{Penlington2020}. To transform the available biomedical information into informed actions that are helpful for the patients, it is important to have systems that automate the task of simplifying the language. Methods in Natural Language Processing were explored to adapt the biomedical language semantically\cite{sulem-etal-2018-simple} and syntactically to automatic plain language at the sentence level \cite{jonnalagadda-etal-2009} and paragraph level \cite{Devaraj-etal-2021}. While some research focused on producing a corpus for facilitating Complex word identification \cite{shardlow-etal-2020-complex}. There is still a shortage of high-quality parallel datasets for training and fine-tuning the models. PLABA was benchmarked using transformer-based models in both zero-shot setting and with fine-tuning. The models for benchmarking PLABA include Text-to-text transfer transformer (T5)-base\cite{raffel2023t5}, Pre-training with extracted gap-sentences for abstractive summarization sequence-to-sequence (PEGASUS)-large\cite{Zhang2019PEGASUSPW}, Bidirectional autoregressive transformer (BART)-base and large\cite{lewis2020bart}, T Zero plus plus (T0PP)\cite{sanh2022multitask}. Recent advancements in the creation and availability of high-quality Pre-Trained Large Language Models have created many opportunities to explore their application in generative tasks such as summarization, reasoning, etc. The availability of the fine-tuned versions of these Pre-Trained language models created to serve the dialog use cases, can be leveraged for any given use case without having to use traditional training techniques to adapt to the dataset\cite{wei2022finetuned}. However, the quality of these examples will affect the quality of the model output\cite{pmlr-v139-zhao21c}. The abstract-level benchmarked results of the PLABA show that using it to further train or fine-tune models could help in automating the adaptation to plain language.

\ifsubfile
\bibliography{mybib}
\fi

\section{Methods}

\subsection{Dataset}
PLABA contains the health-related questions asked by the consumers of MedlinePlus. These questions were filtered based on their frequency and relevance and 75 of them were selected. The focus and question type of each of these questions were sourced by an expert as the keywords to retrieve the 10 most relevant abstracts from PubMed per question, resulting in 750 abstracts in total. Each of these retrieved PubMed abstracts was manually adapted at the sentence level to plain language by at least one annotator. With this, a total of 921 adaptations were created for 750 abstracts. The manual adaptations were created by the annotators following the adaptation guidelines. The adaptation was done by considering the targeted audience's education level as K8. The adaptation guidelines allowed the annotators to drop the complex sentences, retain the simple sentences that need no further simplification as is in the adaptation, splitting the sentences when required, which would allow one or more sentences in the adaptation for each sentence in the abstract. Other guidelines included expansion of abbreviations, omitting statistical figures like p-values, explaining medical jargon with parentheses or nonrestrictive clauses in the first mention as needed, for example, when an abstract introduces a medicine  ``Duloxetine'', adapting it like ``Duloxetine (a common antidepressant)''
\begin{center}
   \begin{table}[!htp]\centering

\begin{tabular}{lcccc}\toprule
&abstract &adaptations \\\midrule
total  &750 &921 \\
total sentences &7612 &9087* \\
avg. no.of sentences &10.15 &10.12 \\
avg. tokens per sentence&22.54 &24.07 \\
\bottomrule
\end{tabular}
\caption{PLABA statistics. * indicates that 235 sentences in the adaptations that are blanks are excluded in this table.} 
\label{data_statistics}
\end{table}  
\end{center}

\subsection{Plain language adaptation using LLMs}
Recent work \cite{Karabacak2023llm, singhal2022large} evaluating LLMs in clinical and biomedical tasks has shown that LLMs are capable of high performance in an array of tasks, including question-answering, information extraction, and summarization. Our experimentation focused on evaluating LLMs for adapting technical language into plain language adaptations. Specifically, we deployed LLMs in both fine-tuned using Parameter Efficient Fine-Tuning (PEFT) \cite{dettmers2023qlora} and in-context learning settings, tuning the models using short instructive sentences to detailed guidelines. We evaluated the models using automatic evaluation metrics and performed human evaluation on a sample of internal test set model generations.

\begin{center}
\begin{figure*}[ht!] 
    \centering
    \frame{\includegraphics[width=5.00in]{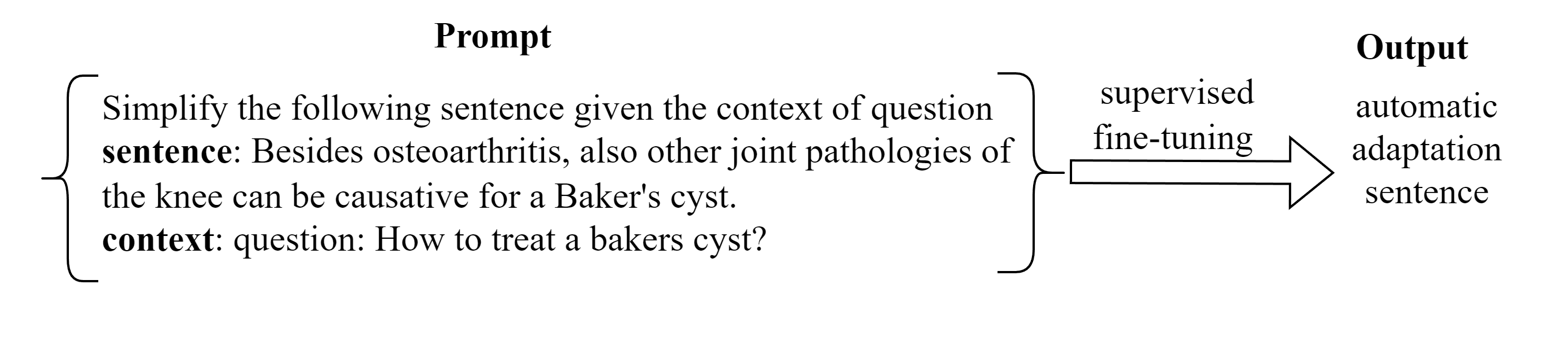}}
    \centering
    \caption{elements of the Prompt structure for Fine-tuning LLaMa-2 models}
    \label{LLaMa-2-prompt}
\end{figure*}
\end{center}

\textbf{Fine-tuning LLMs using PEFT }
Prior work \cite{attal2022dataset} in PLABA demonstrated the performance of fine-tuned T5 \cite{raffel2023t5}  model. We evaluated T5 to replicate the results from prior work but at the sentence level. Additionally, to evaluate the performance of larger generative LLMs, we fine-tuned LLaMa2(13B-chat and 70B-chat) \cite{touvron2023llama} models using Parameter-Efficient Fine-Tuning (PEFT). PEFT allows for fine-tuning the Large Pre-trained models efficiently by reducing the trainable parameters that are suitable for low-resource settings. LLaMa-2 chat models are instruction-tuned using the second-generation LLaMa models from Meta. Two types of fine-tuning are supported in these models - Supervised Fine-Tuning(SFT) and Reinforcement Learning with Human Feedback (RLHF). Further Fine-tuning the instruction-tuned models on a high-quality dataset has proven their efficiency \cite{gupta2023instruction,wei2022finetuned}. We trained the LLaMa2 chat models using SFT by providing instructions in the form of a short instructive sentence, including the Medline search query ( `Simplify the following sentence given the context of the question $<$consumer question$>$'). Our additional experiments training the LLaMa2 70B-chat model and including abstract-specific inter-sentence context for a target sentence did not significantly improve performance. Figure\ref{LLaMa-2-prompt} has the prompt structure used for the LLaMa model.

\begin{center}
\begin{figure*}[ht!] 
    \centering
    \frame{\includegraphics[width=5.00in]{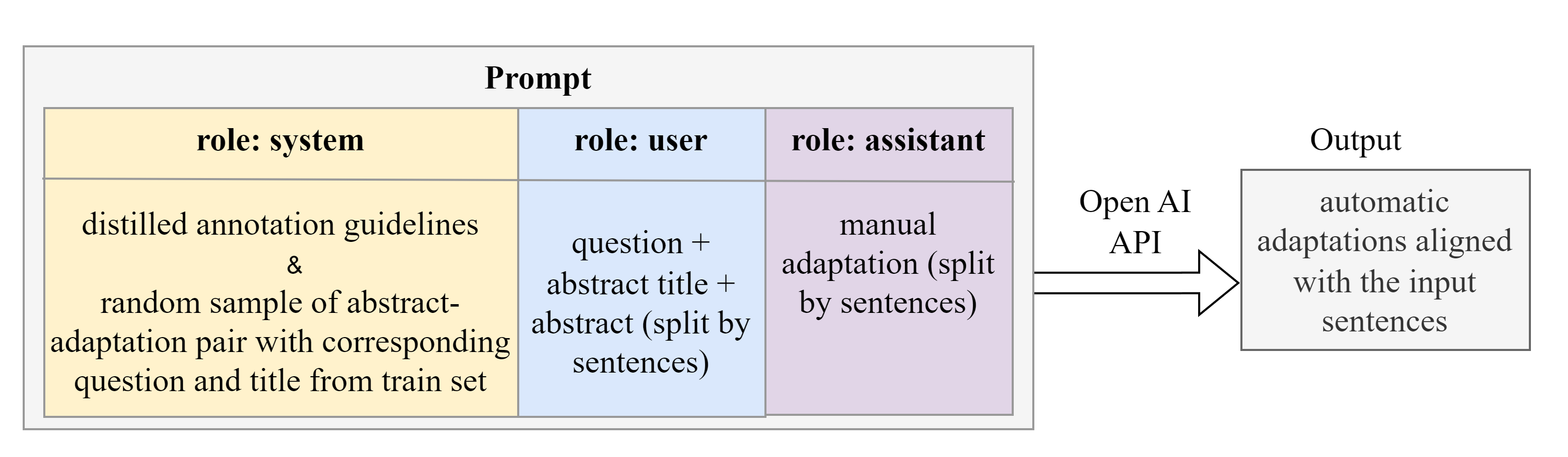}}
    \caption{Prompt structure for the ICL using GPT-4 and Fine-tuning GPT-3.5}
    \label{GPT-J-prompt}
\end{figure*}
\end{center}

\textbf{In-context learning(ICL)  using GPT-J}
We adapt GPT-3.5 \cite{ye2023gpt3.5and3,brown2020gpt3} and GPT-4\cite{openai2023gpt4} for PLABA using in-context learning and fine-tuning. We experimented with a variety of in-context learning strategies for GPT-4, ranging from short prompt instructions to detailed guidelines. Our final GPT-4-based model included a detailed distilled version of the PLABA annotation guideline \cite{PLABA-anno-guide} with one training example that illustrates a majority of annotation guideline instructions. We used the openAI's training application program interface (API) to train GPT-3.5 on the same instruction used for GPT-4, albeit with all abstracts in the training set. Figure \ref{GPT-J-prompt} shows the details of the prompt used for GPT-4 and GPT-3.5 

\subsection{Experimental setup}
We used the PLABA dataset containing 750 abstracts and 921 adaptations for training and fine-tuning our models. 171 abstracts are adapted more than once, in order to create abstract-adaptation sentence pairs for training, we repeat the sentences in the abstract as many times as the number of corresponding adaptations creating a total of 9216 abstract-adaptation sentence pairs. We split the PLABA dataset into \textit{internal} training (70\%), validation (15\%), and test sets (15\%). The internal training and validation sets were used to train and assess the model during the training phase. The internal test set was used to comparatively evaluate all our models to select the most promising models based on automatic and human evaluation. We used internal training, validation and test sets to fine-tune GPT-3.5, LLaMa2 (13b-chat) and T5-Large. GPT-4 on the other hand, used only the internal test set and a single example from train set as we didn't fine-tune the GPT-4 model. 
For fine-tuning T5-Large model, we used batch sizes 8 and 16 with a max sequence length of 256 and 512 and trained for 20 epochs. However, the automatic evaluation scores for all our T5 fine-tuning experiments didn't differ much from each other.
We fine-tuned Llama2 chat models in 4-bit precision using the Low-Rank Adaptation (LoRA), a type of Parameter Efficient Technique for fine-tuning pre-trained language models. We used a LoRA rank set to 16 that allowed training only 0.05\% of the entire trainable parameters.

\subsection{Evaluation}
Automatic evaluation metrics such as BLEU or ROUGE are most commonly used in machine translation. However, there is a need for human judgment to assess the quality and accuracy of the Plain language adaptations because the automatic metrics operate on the n-gram level and are not efficient in capturing the semantics of the generated text. 
\subsubsection{Evaluation on internal test set}
We evaluate the automatically generated adaptations using automatic and human evaluation methods--
(1)Automatic evaluation metrics include- Recall-Oriented Understudy for Gisting Evaluation (ROUGE)\cite{lin-2004-rouge}, BLEU \cite{papineni-etal-2002-bleu}, and SARI \cite{sun-etal-2021-document}. BLEU captures the percentage of words in the automatic adaptations that appear in the given manual adaptations, although it doesn't measure recall directly, it has a brevity penalty to account for shorter auto adaptations. ROUGE on the other hand is a recall-based metric that measures the percentage of words in the manual adaptations that appear in the automatic adaptations. While BLEU and ROUGE compare only the automatic adaptations against the reference text i.e., manual adaptations, SARI considers the abstracts to ensure that the original meaning is preserved and it measures the quality of the words that are added, deleted, and retained from the original text. 
(2) Human evaluation-- sentence simplicity, completeness, accuracy, fluency, and faithfulness criteria. For human evaluation, the lead authors randomly chose five Medline consumer questions and two abstract adaptations per question (approx.100 sentences per model) and evaluated all our fine-tuned and in-context learning models using a simple preference ranking mechanism.  Our fine-tuned models were then trained on the entire PLABA corpus and predicted on the \textit{external} test set shared by the organizers.
\subsubsection{External evaluation on the task test set}
Each participating team was allowed up to three submissions for the final automatic evaluation. Organizers used sentence-level SARI scores to automatically evaluate our submissions on the task test set. For the manual evaluation, task organizers asked the teams to rank each of their three submissions in their preferred order of evaluation, and then they considered the top-ranked submission for the manual evaluation. 
The evaluation was primarily based on two axes\cite{PLABA-evaluation-notes} - (1) Simplicity axis comprising of sentence simplicity, term simplicity, term accuracy, and fluency measured at a sentence level (2) Accuracy axis comprising of faithfulness and completeness measured for up to 3 sentences that are most relevant to the question determined by the annotators. In addition, average scores for each of these axes were shared by the organizers. Table \ref{eval_axes} has a detailed description of the evaluation metrics. 
\begin{center}
    \begin{table}[!htp]\centering
\scriptsize
\begin{tabular}{lll}\toprule
Evaluation axis &Evaluation criteria &Description of Evaluation criteria \\\midrule
&\textbf{Sentence simplicity} &Are long, complex sentences appropriately split? \\
Simplicty axis (determined  &\textbf{Term simplicity} &Are expert terms in the source replaced with alternatives or explained in the output? \\
for each sentence) &\textbf{Term accuracy} &Are substitutions and explanations of expert terms accurate? \\
&\textbf{Fluency} &Does the output follow grammatical rules and read smoothly? \\
\midrule
Accuracy axis (determined for &\textbf{Completeness} &How much of the source information does the output provide? \\
up to 3 most relevant sentences) &\textbf{Faithfulness} &Do points made in the output match those of the source? \\
\bottomrule
\end{tabular}
\caption{ Human Evaluation criteria } 
\label{eval_axes}
\end{table} 
\end{center}

\ifsubfile
\bibliography{mybib}
\fi

\section{Results and Discussion}

\subsection{Automatic evaluation}
We evaluated our models on the automatic metrics-- BLEU, ROUGE-1,2 (based on 1 and 2 grams), ROUGE-L(based on longest shared subsequence between candidate and reference), and SARI scores. While the fine-tuned models T5 and LLaMa show higher performance than the GPT-J models in automatic evaluation, upon manual evaluation of these results, we observed that T5 and LLaMa2 tend to frequently repeat the abstract sentence as is with minimal simplifications, leading to higher ROUGE and BLEU scores. GPT models, on the other hand, captured the goal of the adaptation task by frequently simplifying complex medical condition-related descriptions (e.g. ``macular degeneration'') into understandable plain language (e.g. ``damage to the central part of the retina'') adaptations. 

\begin{center}
    \begin{table}[!htp]\centering
\begin{tabular}{llcccc}\toprule
&& & & &\textsuperscript{1}GPT4- \\
& &&\textsuperscript{3}LLaMa2 & &guide+ \\
&&T5-large &(13B)  & \textsuperscript{2}GPT3.5 & {\small oneshot}\\\midrule
&BLEU &0.232 &\textbf{0.258} &0.176 &0.176 \\
&ROUGE-1 &0.536 &\textbf{0.568} &0.481 &0.503 \\
Internal&ROUGE-2 &0.341 &\textbf{0.357} &0.257 &0.261 \\
&ROUGE-L &0.488 &\textbf{0.522} &0.425 &0.448 \\
&SARI &0.398 &\textbf{0.424} &0.388 &0.396\\
\midrule
External& SARI &0.398 &0.383 &\textbf{0.400} &0.395\\
& & & &Top score &\textbf{0.446}\\
 & & & &Median score &0.395\\

\bottomrule

\end{tabular}
\caption{Automatic evaluation on the internal test set and task test set (external). 1, 2, and 3 denote the ranking of our systems based on internal human evaluation. Final automatic evaluation scores (sentence level SARI score) of all our three submissions. Comparison of our rank1 system's score with the top and median scores of other rank1 systems} 
\label{perf}
\end{table}
\end{center}

For the evaluation of the task test set, all the participating teams were allowed to submit up to three submissions. We submitted our runs from GPT-4, GPT-3.5, and LLaMa-2 models. All these systems were automatically scored at the sentence level using the SARI score metric by the task organizers. Table \ref{perf} contains the performance of our LLM-based fine-tuned and in-context learning models on the internal test set as well as on the task test set and top and median scores of the other rank1 submissions for comparison. 
 
\subsection{Human evaluation}

We performed an internal human evaluation on the internal test set to rank our submissions. GPT-4 ranked first overall with first preference in 9 out of 10 randomly selected abstracts, followed by the fine-tuned GPT-3.5 and LLaMa2 model-generated adaptations. The performance of our GPT-4 model was remarkably higher in internal human evaluation, showing significantly concise and faithfulness in adaptations compared to GPT-3.5 and LLaMa2- whose generations were longer and verbose or were simply repetitions of the complex abstract sentence without any simplification. Among our three submissions, we chose the results from our GPT-4 model as rank 1 based on our internal human evaluation. 

To score the systems, organizers chose one abstract per question for all 40 questions in the evaluation test set. Each of these summaries was scored by annotators on a score of -1, 0, 1 which were then transformed into a scale of 1-100. So the evaluation scores we received were on a 1-100 scale. 
\begin{center}
\begin{table}[!htp]\centering
\begin{tabular}{llcccc}\toprule
& &Ours &Top score &Median score &Rank \\\midrule
\multirow{5}{*}{Simplicity axis} &Sentence simplicity &91.63 &94.33 &91.45 &3 \\
&Term simplicity &\textbf{91.74} &91.74 &81.94 &1 \\
&Term accuracy &88.26 &94.11 &87.5 &3 \\
&Fluency &93.49 &95.25 &92.96 &2 \\
&Simplicty axis avg. &\textbf{91.28} &91.28 &88.86 &1 \\
\midrule
\multirow{3}{*}{Accuracy axis} &Completeness &94.44 &95.73 &90.17 &3 \\
&Faithfulness &90.6 &94.44 &88.46 &3 \\
&Accuracy axis avg. &92.52 &94.87 &89.32 &3 \\
\bottomrule
\end{tabular}
\caption{Final Human Evaluation scores of our system and comparison} 
\label{human_eval_scores}
\end{table} 
\end{center}
There were a total of 7 participating teams and all participants' top-ranked systems were manually evaluated by the expert(s). Simplicity axes evaluation was performed on a total of 430 sentences from the sampled abstracts for each question, while the accuracy axis evaluation was done on a total of 117 sentences. Table \ref{human_eval_scores} provides the final human evaluation results for our top-ranked submission. In addition, for each evaluation criterion, we present the top score, median score, and our ranking among all the 7 participating teams for comparison.

\ifsubfile
\bibliography{mybib}
\fi

\section{Conclusions}

In this work, we propose multiple state-of-the-art LLM-based systems, including T5-large, LLaMa2, GPT-3.5, and GPT4, for plain language adaptation of biomedical abstracts. We trained our LLMs using (1) fine-tuning and (2) in-context learning methods to elicit adaptations. We used both automatic metrics and human evaluation (ranking) to evaluate the model-generated adaptations. While LLaMa2 performed best in automatic evaluation, our best-performing system in human evaluation was the annotation-guideline-instructed GPT-4-based one-shot in-context learning model. Our experiments show that LLMs hold a lot of promise in making complex language in biomedical abstracts more accessible to the consumer audience.

\ifsubfile
\bibliography{mybib}
\fi

\section{Acknowledgements}
This work was supported in part by the National Institutes of Health, National Library of Medicine (NLM) (Grant numbers T15LM007442, R01 CA248422-01A1, R15 LM013209). The content is solely the responsibility of the authors and does not necessarily represent the official views of the National Institutes of Health.

\bibliography{mybib}
\end{document}